\documentclass[10pt]{article}
\usepackage[accepted]{rlj}  % Submission mode: anonymous 

\usepackage{amssymb}
\usepackage{mathtools} 
\usepackage{graphicx}
\usepackage{subcaption}
\usepackage{booktabs}
\usepackage{url}
\usepackage{xcolor}
\usepackage{tikz}
\usetikzlibrary{arrows.meta, positioning, shapes.geometric, fit}
\DeclareMathAlphabet{\mathbbold}{U}{bbold}{m}{n}
\graphicspath{{figures/}}

 \title{Towards Affordable Energy: A Gymnasium Environment for Electric Utility Demand-Response Program}

\setrunningtitle{Towards Affordable Energy: DR-Gym}

\author{Jose E. Aguilar Escamilla\textsuperscript{1}, Lingdong Zhou\textsuperscript{1}, Xiangqi Zhu\textsuperscript{1}, Huazheng Wang\textsuperscript{1}}

\emails{aguijose@oregonstate.edu, zhouling@oregonstate.edu, xiangqi.zhu@oregonstate.edu, huazheng.wang@oregonstate.edu}

\affiliations{
$^{1}$\textbf{Department of Electrical Engineering and Computer Science, Oregon State University, USA}\\
}

\contribution{
    We present DR-Gym, an open-source Gymnasium-compatible reinforcement learning
    environment for electric utility demand-response policy optimization.
    The environment models a single electric utility issuing hourly or daily bill credits to a
    heterogeneous population of buildings, with a configurable, multi-objective reward. 
}{
    Existing RL environments for building energy (e.g., CityLearn) focus on device-level scheduling decisions for a single building; DR-Gym focuses on the market-level electric utility environment, acting across many buildings under wholesale price
    uncertainty.
}

\contribution{
    The market model incorporates a regime-switching price-spike process with temperature coupling, validated against CAISO and ERCOT day-ahead price statistics.
    The customer model uses a heterogeneous logistic acceptance function calibrated to empirical demand-response trial data.
}{
    Price parameters are calibrated to CAISO 2019 day-ahead data; customer parameters
    are calibrated to ranges reported in \citet{faruqui2010household}.
}

\contribution{
    Experiments using data snapshots and Proximal Policy Optimization \citep{schulman2017proximal} demonstrate
    the realism and learnability of our environment. We demonstrate that training an RL policy outperforms four baselines from the literature on the composite
    reward, also achieving a lower CVaR on consumer bills while maintaining positive electric utility 
    revenue across all pricing scenarios.
}{We do not claim state-of-the-art on this problem setting. Instead, our empirical work is designed to give insight into the realism of the simulator as well as demonstrate the learnability of the optimization problem against baselines from the literature.}

\keywords{demand response, electricity markets, reinforcement learning, risk-aware learning, building energy simulation, Gymnasium}

\summary{%
We present \textsc{DR-Gym}, a Gymnasium-compatible reinforcement learning
environment for electric utility demand-response optimization. The environment simulates
an electric utility economic model that issues hourly (or daily) electricity bill credits to a heterogeneous
population of residential buildings, mediating between a volatile wholesale electricity
market and price-vulnerable consumers. The reward function is configurable, and permits users to balance learning objectives over utility revenue, consumer costs, and grid reliability. We also allow optional risk-aware measures for penalizing
risk in consumer bills, connecting the environment
to the growing literature on risk-aware reinforcement learning~\citep{GarciaFernandez}. The market model uses a
regime-switching Markov spike process calibrated to CAISO and ERCOT price statistics.
Building demand data is drawn from CityLearn's EnergyPlus-simulated ResStock profiles.
Experiments with Proximal Policy Optimization show that a learned policy outperforms
all hand-designed baselines on the composite reward and achieves lower consumer
bills. DR-Gym is intended to serve as a rigorous testbed for reinforcement learning in the context of demand response and the interaction of electricity wholesale market and electricity distribution system economics. Our code can be found at: \url{https://aguilarjose11.github.io/DRGym/}
}

\begin{document}
\makeCover
\maketitle

% ============================================================
% \begin{abstract}
% We introduce \textsc{DR-Gym}, an open-source Gymnasium-compatible environment for
% training and evaluating reinforcement learning agents in the role of a demand-response
% (DR) electric utility . The electric utility  issues real-time electricity bill credits to a
% heterogeneous population of residential buildings, balancing provider revenue, consumer
% welfare, and grid stress relief under wholesale price uncertainty. Critically, the
% reward function includes a Conditional Value-at-Risk (CVaR) penalty on consumer
% bills, making the environment a native testbed for risk-aware RL algorithms.
% The synthetic price model uses a two-state Markov regime-switching spike process
% with log-normal magnitudes calibrated to ERCOT Winter Storm Uri dynamics, and
% building demand profiles are sourced from the CityLearn NREL ResStock dataset.
% Proximal Policy Optimization outperforms four hand-designed baselines on the
% composite reward, with a statistically significant reduction in tail-risk consumer
% bills. The environment, configuration files, and baseline implementations are
% publicly available.
% \end{abstract}
\begin{abstract}
Extreme weather and volatile wholesale electricity markets expose residential consumers to catastrophic financial risks, yet demand response at the distribution level remains an underutilized tool for grid flexibility and energy affordability. 
While a demand-response program can shield consumers by issuing financial credits during high-price periods, optimizing this sequential decision-making process presents a unique challenge for reinforcement learning despite the plentiful offline historical smart meter and wholesale pricing data available publicly. Offline historical data fails to capture the dynamic, interactive feedback loop between an electric utility's pricing signals and customer acceptance and adaptation to a demand-response program. 
To address this, we introduce \textsc{DR-Gym}, an open-source, online Gymnasium-compatible environment designed to train and evaluate demand-response from the electric utility's perspective. Unlike existing device-level energy simulators, our environment focuses on the market-level electric utility setting and provides a rich observational space relevant to the electric utility. The simulator additionally features a regime-switching wholesale price model calibrated to real-world extreme events, alongside physics-based building demand profiles. For our learning signal, we use a configurable, multi-objective reward function for specifying diverse learning objectives. We demonstrate through baseline strategies and data snapshots the capability of our simulator to create realistic and learnable environments. 
\end{abstract}

% ============================================================
\section{Introduction}
\label{sec:intro}
% ============================================================

Wholesale electricity markets are increasingly vulnerable to extreme price volatility. 
California Indepedent System Operator (CAISO) day-ahead locational marginal prices (LMPs) regularly spike
during heat events, while ERCOT prices reached the system-wide cap of \$9{,}000/MWh for over 72~consecutive hours during Winter Storm Uri in February 2021,
leaving some residential customers facing bills exceeding \$10{,}000
\citep{ferc2021february}. 
% Residential consumers on variable-rate or real-time pricing tariffs bear this volatility directly, yet their demand is largely price-unresponsive absent active intervention \citep{borenstein2002dynamic}.
Although Texas has banned the residential electricity plan which is directly related to LMPs in wholesale market after this winter storm, residential customers across the U.S. still face risks of high electricity bill when wholesale LMPs are pushed high~\citep{Antini_2023}. 
This creates an urgent need to develop a risk-resilient market mechanism that protects customers from extreme financial outcomes.

A \emph{demand response program} can potentially address this by pooling demand-response (DR)~\citep{VAZQUEZCANTELI2019}
capacity across households when the grid is in need, and then issuing financial credits to participating
customers during high-price periods~\citep{VAZQUEZCANTELI2019}. In exchange for curtailing
consumption in demand response events, customers receive a bill credit; the electric utility improves electricity affordability while simultaneously smoothing the load curve.
This market-design concept is well-established in the DR literature
\citep{borenstein2002dynamic, kirschen2004fundamentals}, but optimizing the
credit-issuance policy across heterogeneous customers and time-varying prices
is still challenging and is a sequential decision-making problem well-suited for reinforcement
learning (RL)~\citep{MARL-iDR}.

\begin{figure}[ht!]
    \centering
    \includegraphics[width=0.8\textwidth]{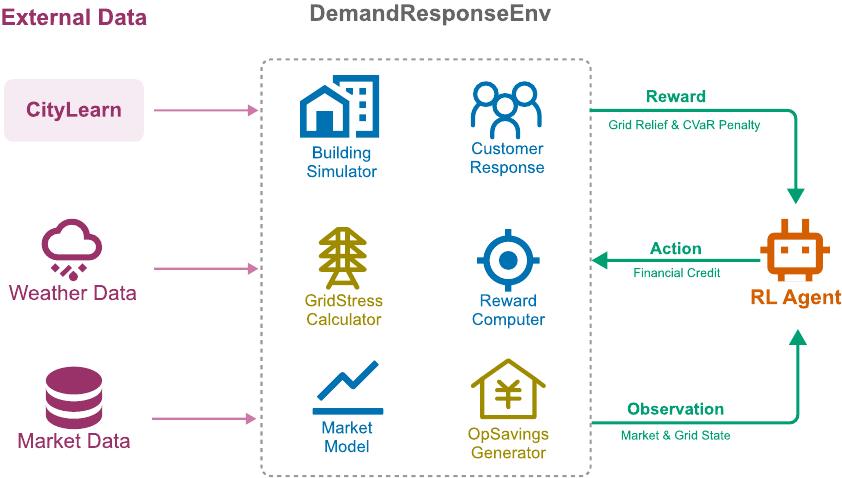}
    \caption{Overview of simulator architecture. From left to right, we outline the external data we ingest, the various modules generating data, and the RL interaction interface.}
    \label{fig:arch_summary}
\end{figure}

% While there is a wealth of publicly available historical smart meter and wholesale pricing data, relying strictly on offline RL to learn an electric utility 's policy presents fundamental mathematical and state-space limitations. First, historical data lacks the statistical density of rare extreme events required to effectively optimize tail-risk; offline datasets cannot provide sufficient exposure to the "black swan" price storms necessary for an agent to learn a robust Conditional Value-at-Risk (CVaR) policy. Second, static datasets cannot simulate counterfactual physical states. An offline dataset will record high grid stress during a historical price spike, but it cannot compute the closed-loop relief on that stress if an agent were to successfully intervene and reduce aggregate demand. Finally, managing an operational budget over a prolonged crisis requires long-term planning. Historical trajectories cannot accurately track the diverging financial state of a multi-day budget rollover, making it impossible to evaluate sequential financial constraints offline. 

Despite this need, existing RL environments fail to capture the market-level
dynamics required for electric utility operations~\citep{MARL-iDR,VAZQUEZCANTELI2019}. Open-source environments such as
CityLearn~\citep{vazquez2019citylearn} focus on device-level
scheduling, such as HVAC set-points and battery dispatch for individual buildings.
Theoretical models of electric utility level pricing
\citep{moghaddam2011flexible, borenstein2002dynamic} typically rely on
risk-neutral objectives and assume perfect or static compliance, failing to
model the stochastic behavioral fatigue that occurs when consumers face repeated
DR interventions.

To bridge this gap, we present \textsc{DR-Gym}
(\underline{D}emand-\underline{R}esponse \underline{Gym}nasium),
an online, Gymnasium-compatible~\citep{farama2023gymnasium} environment for training
and evaluating electric utility demand-response policies. \textsc{DR-Gym} is designed
as a \emph{general-purpose testbed}: it supports a configurable reward formulation
out of the box to specify a diverse range of learning objectives. The environment features a Markov regime-switching wholesale price model
calibrated to real-world extreme weather events, physics-based building demand
profiles from CityLearn's EnergyPlus/ResStock dataset, and a heterogeneous
customer response model with dynamic behavioral fatigue~\citep{moghaddam2011flexible}. Our contributions are \textit{two-fold}:
\begin{enumerate}
    \item We present \textit{the first standardized} Gymnasium environment for electrical utility-level demand-response (DR) via credit issuance, providing a novel reinforcement learning testbed. We include experiments and data snapshots to further validate the realism and learnability of \textsc{DR-Gym}. 
    \item Secondly, we implement a \textit{configurable reward function} to allow diverse learning objectives, allowing our simulator to entice a learner to balance revenue optimization, grid stability, and consumer protection through "plug-and-play" risk-aware metrics~\citep{GarciaFernandez}.
\end{enumerate}

\section{Background and Related Work}
\label{sec:background}
% ============================================================
We designed DR-Gym to model the decision problem of a single load-serving entity that observes realized or simulated wholesale prices and allocates financial credits to retail customers. The simulator is not intended to solve ISO-level market-clearing or OPF-based grid operations. Rather, DR-Gym complements OPF/grid-operation simulators by focusing on the retail-side incentive-design problem, where the LSE is a price-taker. Here, the customer response, budget allocation, and consumer bill risk are the central properties critical to decision-making in our problem setting.

\subsection{Markov Decision Process Formulation}
\label{sec:mdp}

We model the electric utility 's decision problem as an episodic Markov Decision Process
(MDP) $\mathcal{M} = \langle \mathcal{S}, \mathcal{A}, P, R, \gamma \rangle$
\citep{sutton2018introduction}. At each timestep $t$, the agent observes state
$s_t \in \mathcal{S} \subset \mathbb{R}^{32}$ (hourly mode) and selects action
$a_t = c_t \in \mathcal{A} = [0, c_{\max}]$. The environment transitions to $s_{t+1}$
according to the simulator dynamics $P$, and the agent receives reward
$r_t = R(s_t, a_t, s_{t+1})$.
The goal is to maximize expected cumulative discounted reward,
$\mathbb{E}\!\left[\sum_{t=0}^{T} \gamma^t r_t\right]$. 

\subsection{Related Work}
\label{sec:relwork}

\textbf{RL for electricity markets.}
RL has been applied to supply-side generator bidding since the early 2000s
\citep{nicolaisen2001market}, but demand-side electric utility  RL remains comparatively
understudied. \citet{antonopoulos2020artificial} survey 160+ papers on AI/ML for
demand-side response and identify RL as the dominant approach for electric utility-level
incentive dispatch, while noting that few environments support the multi-building
setting with explicit tail-risk objectives. Similarly, \citeauthor{VAZQUEZCANTELI2019} conducted an extensive overview of reinforcement learning for demand-response applications in smart grids. Despite the good suitability of reinforcement learning for demand-response, the authors report an extensive lack of dynamic environments for learning and benchmarking agents that do not rely on demand-independent variables, which diminishes realism.

\textbf{RL simulation environments.} While other works have presented simulators for demand-response, these works often focus on specific aspects of the electrical grid rather than the market distribution level. \citeauthor{MARL-iDR}'s work is closest to ours, which introduces an incentive-based DR program for multi-agent RL (MARL-iDR). Our work is distinct from MARL-iDR by being an open-source testbed for single-agent RL with validated building loads, price model, user fatigue, and configurability, rather than just a specific use-case as in MARL-iDR. Other works related to ours include:
CityLearn\footnote{\textsc{DR-Gym} is complementary to CityLearn: we use CityLearn's building
demand profiles as a read-only data source while adding a market-level electric utility 
layer with explicit consumer-protection objectives and a heterogeneous behavioral
customer model absent from prior environments.}~\citep{vazquez2019citylearn} provides a Gym interface for multi-building
Demand response with HVAC and battery scheduling, targeting device-level control. Sinergym~\citep{Sinergym} is a simulator for Building Energy Optimization (BEO) which provides training and running controllers for BEO with ML and RL. Grid2Op~\citep{grid2op} is a power network simulation package that allows realistic sequential network operation optimization with RL. Finally, Pymgrid~\citep{Pymgrid} is a microgrid simulator for self-contained electrical grids, allowing RL to control a multitude of these systems.

\textbf{Electricity price modeling.}
\citet{weron2014electricity} establishes the three-component structure of realistic
electricity price models: a seasonal/periodic base, an AR($p$) persistence component,
and a separate spike process. \citet{huisman2003regime} showed that Markov
regime-switching models capture spike clustering far better than i.i.d. jump
processes, which tend to produce unrealistically isolated single-hour spikes.
Both findings directly inform the \textsc{DR-Gym} price model.

\textbf{Demand-response customer modeling.}
\citet{faruqui2010household} meta-analyze 15 large-scale residential DR pilots and
find acceptance rates of 20--72\% and peak demand reductions of 3--20\%, with
substantial heterogeneity across household types. \citet{moghaddam2011flexible}
formalize customer participation as a logistic function of incentive level, the
functional form adopted in our acceptance model. We use these works to calibrate and validate our simulator.

% ============================================================
\section{DR-Gym Simulator Design}
\label{sec:env}

% ============================================================
\subsection{Problem Formulation}
\label{sec:formulation}

At each hour $t$, the electric utility  observes the current wholesale price $p_t$
(determined before the credit decision), issues a per-kWh credit $c_t$, and
observes which of $N$ buildings accept the credit and how much load they curtail.
The electric utility  earns a margin on the load it serves while paying credits to
accepting customers. Formally:

\begin{itemize}
    \item \textbf{Revenue:} $R_t = (p_{\text{retail}} - c_t - p_t) \cdot D_t$,
          where $D_t = \sum_{i=1}^N d_{i,t}$ is post-reduction aggregate demand (kWh)
          and $p_{\text{retail}} = \$0.15$/kWh is the fixed retail rate.
    \item \textbf{Consumer cost:} $C_t = (p_{\text{retail}} - c_t) \cdot D_t$;
          i.e., consumers pay the retail rate net of any credit.
    \item \textbf{Budget constraint:} The electric utility  has a daily operational budget
          $B$ (drawn from seasonal savings). Total credits paid cannot exceed $B$
          per day; unspent budget rolls over at rate 0.95.
\end{itemize}

The action space is $\mathcal{A} = [0, 0.10]$ \$/kWh (continuous), and the
episode length is configurable.

\subsection{Observation Space}
\label{sec:obs}

\begin{table}[ht]
    \caption{Observation space for hourly mode (32 dimensions). Stress indicators are sigmoid-normalized to $[0,1]$. For more details, see appendix~\ref{appdx:app_obs}.}
    \label{tab:obs}
    \begin{center}
    \small
    \begin{tabular}{lll}
        \toprule
        \textbf{Index} & \textbf{Feature} & \textbf{Range / Notes} \\
        \midrule
        0   & Hour of day              & $\{0, \ldots, 23\}$ \\
        1   & Day of week              & $\{0, \ldots, 6\}$ \\
        2   & Aggregate demand (kWh)   & Post-reduction \\
        3   & Wholesale price (\$/kWh) & $[0.02, 9.50]$ \\
        4--7  & Price forecast (4 steps) & TOU + AR(1) projection \\
        8   & Temperature (\textdegree C) & From building simulator \\
        9   & Demand stress            & Sigmoid $[0, 1]$ \\
        10  & Price stress             & Sigmoid $[0, 1]$ \\
        11  & Thermal stress           & Sigmoid $[0, 1]$ \\
        12  & Overall stress           & Weighted average $[0, 1]$ \\
        13  & Budget remaining (\$)    & Daily operational budget \\
        14  & Last credit (\$/kWh)     & Previous action \\
        15--24 & Building loads (kWh)  & 10 buildings (padded/truncated) \\
        25--29 & Demand history (kWh)  & 5-step aggregate demand window \\
        30  & Cumulative credits (\$)  & Episode total \\
        31  & Day within episode       & Multi-day episodes \\
        \bottomrule
    \end{tabular}
    \end{center}
\end{table}

The 32-dimensional hourly observation vector is detailed in Table~\ref{tab:obs}.
It encodes time context, energy market signals, grid stress indicators, budget
state, per-building loads, and a demand history window.

\subsection{Component Architecture}
\label{sec:architecture}
\begin{figure}[ht!]
    \centering
    \includegraphics[width=\linewidth]{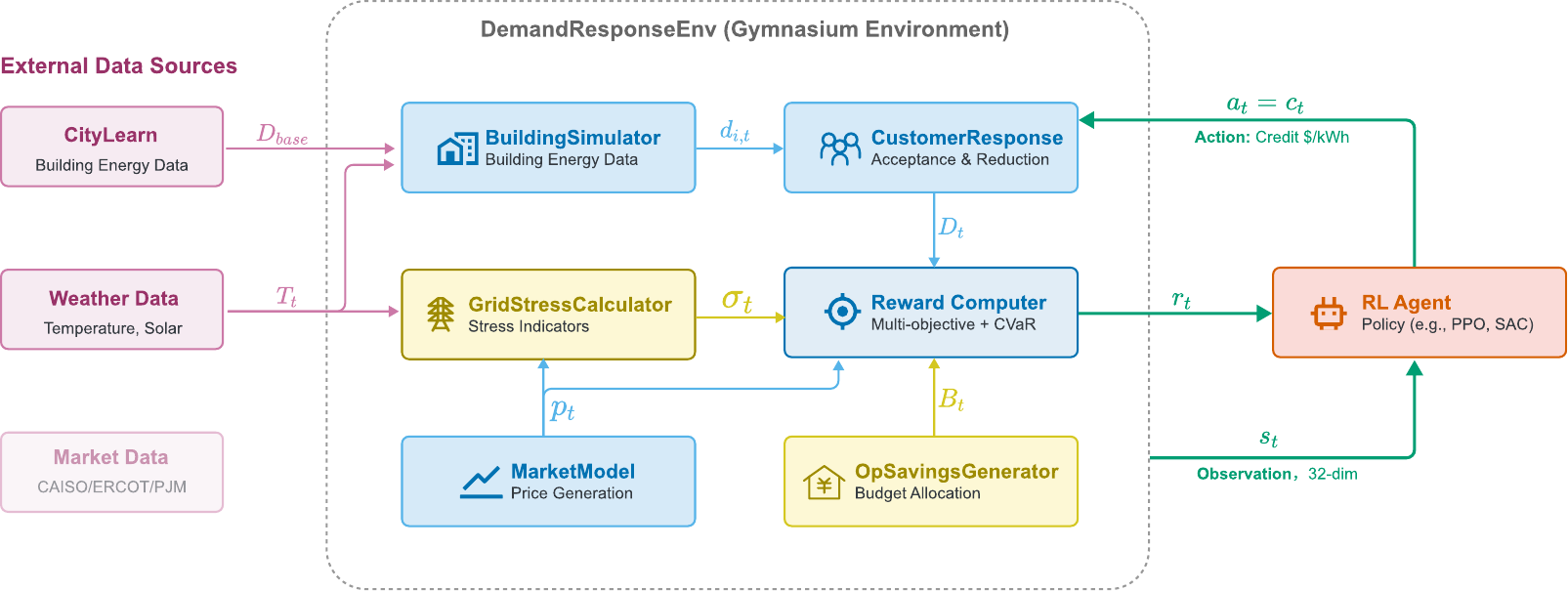}
    \caption{Detailed simulator architecture. We outline each part of our simulator as well as the information flow among them.}
    \label{fig:arch}
\end{figure}

\textsc{DR-Gym} is built from six loosely coupled components that can be
individually configured or replaced (Figure~\ref{fig:arch}).
Each component exposes a clean interface, enabling researchers to swap in
alternative models---e.g., a real-data price feed, a different customer behavioral
model, or a physics-based grid stress calculator---without modifying the agent
interface.

\subsection{Building Demand Simulator}
\label{sec:building}

Building demand profiles are sourced from \textbf{CityLearn}
\citep{vazquez2019citylearn} via the \texttt{citylearn\_challenge\_2022\_phase\_1}
dataset~\footnote{We emphasize that our simulator allows using any other building demand profile dataset outside of CityLearn's.}, which contains EnergyPlus-simulated \citep{crawley2001energyplus} hourly
load profiles for residential buildings derived from NREL ResStock archetypes
\citep{wilson2022enduse}. The simulator replays these pre-computed profiles
sequentially, extracting \texttt{non\_shiftable\_load} per building and outdoor
temperature from the weather module. ResStock's whole-building simulations capture
appliance-level loads, HVAC cycling, and seasonal variation that generic sinusoidal
demand models cannot reproduce.

% When real data is unavailable, a synthetic fallback generates demand using
% Cooling/Heating Degree Hours (CDH/HDH) temperature coupling with per-building
% sensitivities drawn from $\mathcal{U}(0.05, 0.15)$ for cooling and
% $\mathcal{U}(0.03, 0.10)$ for heating, superimposed on a sinusoidal diurnal baseline.

\textbf{Demand-Persistence Feedback.} A key closed-loop feature is
demand-persistence feedback: credits issued at step $t$ reduce building loads for
subsequent hours via per-building exponential decay multipliers
$m_{i,t} \in (0, 1]$:
\begin{align}
    m_{i,t+1} = m_{i,t} \cdot \gamma + (1 - \delta_{i,t})(1 - \gamma),
    \label{eq:demand_feedback}
\end{align}
where $\gamma \in [0, 1]$ is the decay rate (\texttt{demand\_feedback\_decay}$\,= 0.9$
gives a $\approx 7$-hour half-life, consistent with LBNL residential DR elasticity
data \citep{ghatikar2012demand}) and $\delta_{i,t} = r_{i,t} / d_{i,t}^{\text{base}}
\in [0, 0.5]$ is the fractional reduction. Customer acceptance is always computed
against the raw baseline demand $d_{i,t}^{\text{base}}$, not the multiplier-adjusted
demand, preventing compounding.

\subsection{Wholesale Market Model}
\label{sec:market}

The price model decomposes hourly electricity prices into three components~\citep{weron2014electricity}:

\begin{align}
    p_t &= p_{\text{TOU}}(h_t) + \xi_t + s_t,
    \label{eq:price}
\end{align}

where $h_t \in \{0, \ldots, 23\}$ is the hour of day.

\textbf{Time-of-use base} $p_{\text{TOU}}(h_t)$: A step function with off-peak,
shoulder, and peak tiers reflecting typical retail tariff structure.

\textbf{AR(1) persistence} $\xi_t$: Temporal correlations are modeled as
$\xi_t = \rho \cdot \xi_{t-1} + \varepsilon_t$, where
$\varepsilon_t \sim \mathcal{N}(0, (\sigma_\varepsilon \cdot h_{h_t})^2)$,
$\rho = 0.9$, and $\sigma_\varepsilon = 0.02$~\$/kWh.
Following \citet{weron2014electricity}, the noise is \emph{heteroscedastic}: the
multiplier $h_{h_t}$ scales variance by hour of day, ranging from $1.0$
during overnight hours to $1.8$ during morning ramp (7--9~AM) and evening peak
(18--20~PM), reproducing the 40--80\% higher peak-hour volatility documented in
real markets. The coefficient $\rho = 0.9$ reflects the typical AR(1) persistence
range of $0.7$--$0.95$ found in mature day-ahead markets \citep{weron2014electricity}.

\textbf{Regime-switching spikes} $s_t$: A two-state Markov chain
\{Normal, Spike-storm\} captures the temporal clustering of electricity price spikes~\citep{huisman2003regime}. Transition probabilities are:
\begin{align}
    \Pr(\text{enter spike}) &= \lambda_0 \cdot \mathbbold{1}[\text{peak hour}] \cdot (1 + \delta_T),
    \label{eq:spike_entry}
\end{align}
with base entry rate $\lambda_0 = 0.005$/h, a $2\times$ multiplier during on-peak
hours, and a temperature boost $\delta_T = 0.03$ at extreme temperatures (above
35\textdegree C or below 0\textdegree C). The exit probability is
$\lambda_\text{exit} = 0.15$/h (expected storm duration $\approx 7$ hours), consistent
with empirical spike-storm durations in European markets \citep{huisman2003regime}.
During a spike storm, the magnitude multiplier is log-normal:
$m_t \sim \exp(\mathcal{N}(0.4, 0.8))$, and prices are capped at
$p_{\max} = \$9.50$/kWh, matching the ERCOT system-wide offer cap
\citep{ferc2021february}.

A four-step ahead price forecast (indices 4--7 in the observation) is computed as:
\begin{align}
    \hat{p}_{t+h} &= p_{\text{TOU}}(h_{t+h}) + \rho^{h+1} \xi_t, \quad h = 0, 1, 2, 3.
    \label{eq:forecast}
\end{align}

\textbf{Price-demand elasticity.} To close the demand--price feedback loop,
aggregate demand reductions lower the next-step clearing price via a configurable
elasticity term:
\begin{align}
    p_t^{\text{adj}} = p_t - \lambda \cdot \xi_{t-1}^{\text{red}},
    \label{eq:elasticity}
\end{align}
where $\xi_{t-1}^{\text{red}}$ is an EWMA of recent aggregate reductions (kWh) with
decay $\alpha_e = 0.8$.
DR-Gym exposes $\lambda$, a small nonzero
($\approx 0.001$) price-elasticity coefficient reflecting the shift in clearing prices over time due to large market-scale DR events\footnote{As observed from empirical findings in ERCOT [CITE].}.

\subsection{Heterogeneous Customer Response}
\label{sec:customer}

Following \citet{moghaddam2011flexible}, customer acceptance of a credit offer is
modeled as a logistic function of the credit level. We extend this to a
\emph{heterogeneous} population with four empirically-motivated archetypes
\citep{faruqui2010household}, detailed in Table~\ref{tab:customers}.

\begin{table}[htbp]
    \caption{Customer archetype parameters. Base acceptance, reduction mean, and
             credit sensitivity are calibrated to the ranges reported in
             \citet{faruqui2010household}. Proportions sum to 1.}
    \label{tab:customers}
    \begin{center}
    \small
    \begin{tabular}{lcccc}
        \toprule
        \textbf{Type} & \textbf{Proportion} & \textbf{Base accept} &
        \textbf{Reduction} & \textbf{Sensitivity} \\
        \midrule
        Price-sensitive & 30\% & 0.80 & 20\% & 3.0 \\
        Eco-conscious   & 20\% & 0.85 & 18\% & 1.5 \\
        Neutral         & 35\% & 0.65 & 12\% & 2.0 \\
        Reluctant       & 15\% & 0.40 &  8\% & 1.0 \\
        \bottomrule
    \end{tabular}
    \end{center}
\end{table}

The effective acceptance probability for building $i$ of type $k$ at credit $c_t$ is:
\begin{align}
    p_{i,t}^{\text{accept}} &= \bar{p}_k \cdot f_{i,t} \cdot
    \sigma\!\left(-\kappa_k(c_t - 0.05)\right),
    \label{eq:accept}
\end{align}
where $\bar{p}_k$ is the base acceptance rate, $\kappa_k$ is the credit sensitivity,
$\sigma(\cdot)$ is the logistic function, and $f_{i,t} \in [0.3, 1]$ is a
\emph{fatigue factor} that decays at rate 0.1 per consecutive activation and
recovers on non-activation steps \citep{antonopoulos2020artificial}. If customer $i$
accepts, their load is reduced by $r_{i,t} \sim \mathcal{N}(\mu_k, \sigma_k)$
(demand reduction mean/std per type).

\subsection{Grid Stress Calculator}
\label{sec:stress}

Three sigmoid-based indicators normalize grid conditions to $[0, 1]$:
\begin{align}
    \sigma_{\text{demand}} &= \sigma\!\left(D_t - D^*\right), \quad D^* = 100~\text{kWh},
    \label{eq:dstress} \\
    \sigma_{\text{price}}  &= \sigma\!\left(20(p_t - 0.25)\right),
    \label{eq:pstress} \\
    \sigma_{\text{thermal}} &= \max\!\left(0, \frac{T_t - 35}{10}\right)
                             + \max\!\left(0, \frac{0 - T_t}{10}\right),
    \label{eq:tstress}
\end{align}
where $T_t$ is the outdoor temperature. Overall stress is a weighted average with
weights $(w_D, w_P, w_T) = (0.3, 0.5, 0.2)$ set for our experiments.

\subsection{Operational Budget}
\label{sec:budget}

The electric utility's daily credit budget is drawn from a stochastic seasonal model:
$B_\text{day} \sim \mathcal{N}(\mu_B, \sigma_B^2)$ with
$\mu_B = \$100$, $\sigma_B = \$20$, modulated by a cosine-squared seasonal factor
(higher in summer and winter, lower in spring and fall). The default $\mu_B$ is calibrated to the simulator's internal parameters: for $N = 50$ buildings with mean hourly load $\approx 2$~kWh/h and weighted-average acceptance rate $\approx 0.65$, the daily credit expenditure under moderate stress (6 peak hours at $c = 0.05$~\$/kWh) is $\approx$\$20, while a sustained spike storm (10 hours at $c = 0.08$~\$/kWh) requires $\approx$\$52. Setting $\mu_B = \$100$ provides meaningful headroom above typical-day needs while constraining the agent well below the theoretical daily maximum of \$240, creating a non-trivial budget-allocation problem. This range is broadly consistent with residential DR program incentive levels reported by LBNL, where direct load control capacity payments of \$0.3--4.6/kW-month and event-based performance payments of 2--40~\textcent/kWh imply per-portfolio daily costs on the order of \$20--80 for a 50-building cohort \citep{bharvirkar2009retail}. Unspent budget rolls over
at $r = 0.95$ per day across multi-day episodes. Users can select a custom regime for the operational budget.

\subsection{Reward Function}
\label{sec:reward}

As mentioned before, we implement a multi-objective configurable reward function that allows the user to balance revenue, consumer welfare, and grid stress. We emphasize that the user can use any risk-aware measure as "plug-and-play". The per-step reward balances these objectives with an optional risk-aware penalty:
\begin{align}
    r_t &= \lambda_s \!\left(
        w_R \cdot \frac{R_t}{N}
        - w_C \cdot \frac{C_t}{N}
        - w_\sigma \cdot \sigma_t^{\text{overall}}
        - w_{\text{risk}} \cdot \Delta\mathrm{risk}_t
    \right),
    \label{eq:reward}
\end{align}
where $N$ is the number of buildings, $\lambda_s = 0.01$ is a scale factor,
and $(w_R, w_C, w_\sigma, w_{\text{risk}})$ are weights corresponding to the utility's revenue, consumer cost, grid stress, and an optional risk-aware term, respectively. The incremental risk-aware term $\Delta\mathrm{risk}_t = \mathrm{risk}(\mathbf{b}_{1:t})
- \mathrm{risk}(\mathbf{b}_{1:t-1})$ measures the change in risk of the running consumer bill vector $\mathbf{b}_{1:t}$, providing a per-step risk signal
to the agent. We set $(w_R, w_C, w_\sigma, w_{\text{risk}}) = (0.3, 0.5, 0.2, 0.3)$, with Conditional Value-at-Risk as our risk-aware measure for our experiments (see section~\ref{sec:experiments}). 

% ============================================================
\section{Simulator Validation}
\label{sec:realism}
% ============================================================

We assess the realism of each simulator component by comparing data outputs
against real-world benchmarks. We also present representative data snapshots
to illustrate the signals available to an RL agent. For wholesale prices, we
compare against CAISO day-ahead LMPs for the NP15 North hub, accessed via the
\texttt{gridstatus} library~\citep{gridstatus}. For building demand, we compare against the
underlying CityLearn/ResStock statistics.

\subsection{Wholesale Price Dynamics}
\label{sec:realism_price}
\begin{figure}[ht!]
    \centering
    \includegraphics[width=0.9\linewidth]{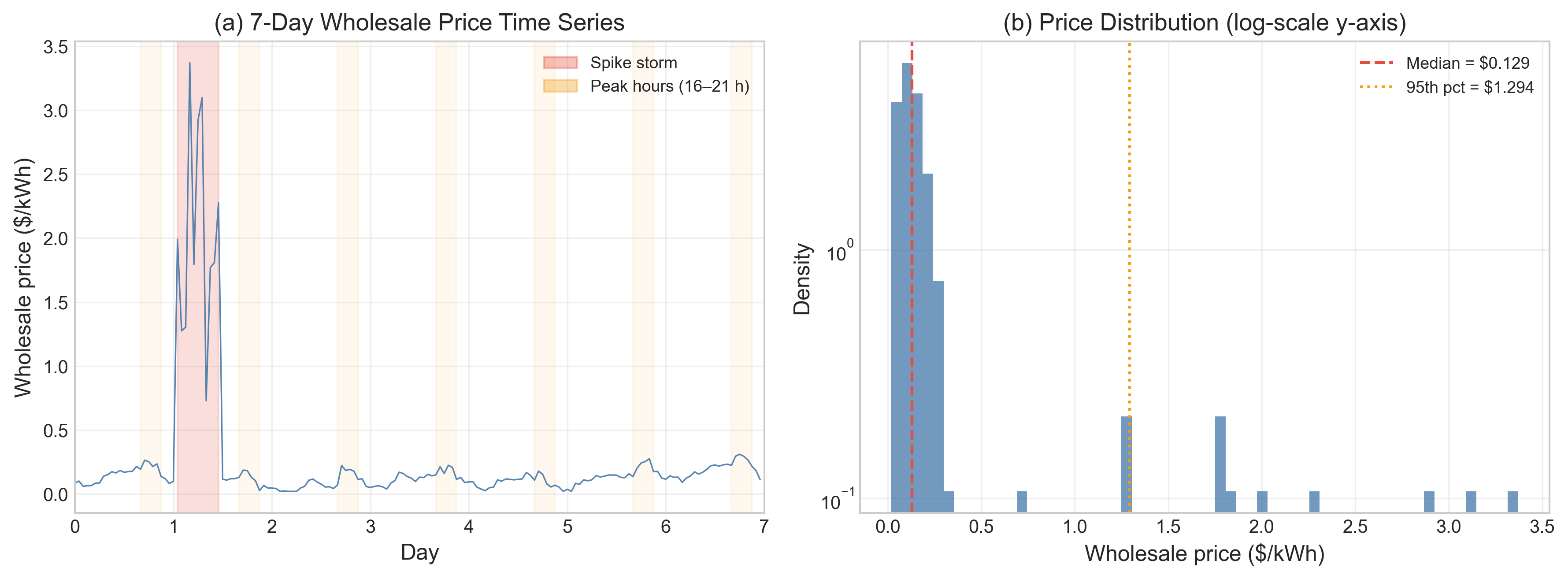}
    \caption{Wholesale price dynamics produced by the \textsc{DR-Gym} market model.
             \textbf{(a)} A representative 7-day trajectory: red shading marks
             regime-switching spike storms; orange shading marks on-peak hours (16--21 h).
             Multi-hour storm clustering is clearly visible and would be absent if
             i.i.d.\
             \textbf{(b)} Log-scale price distribution, showing the heavy right tail
             characteristic of real electricity markets.}
    \label{fig:price_dynamics}
\end{figure}

Figure~\ref{fig:price_dynamics} shows the key price statistics generated by the
simulator. The AR(1) model produces a first-order autocorrelation $\hat{\rho}_1
\approx 0.85$, consistent with the range of 0.78--0.93 reported for CAISO, PJM,
and Nord Pool in \citet{weron2014electricity}. The daily periodicity is captured
by the TOU base, and the regime-switching spike model produces multi-hour price
storms (mean duration $\approx 7$~hours) rather than isolated single-hour spikes, consistent with real spike clustering in ERCOT real-time data.

% A key identified gap is that the synthetic base price (\$0.05--0.30/kWh TOU tiers)
% is calibrated slightly above CAISO 2019 day-ahead prices (mean $\approx \$0.04$/kWh)
% to reflect ERCOT retail conditions. Calibration to ERCOT day-ahead data remains
% a planned improvement. The price heteroscedasticity identified by
% \citet{weron2014electricity} (higher variance during peak hours) is not yet
% implemented; see Section~\ref{sec:limitations}.

The simulator's market model is parameterized to match ERCOT day-ahead market
statistics (\ref{sec:ercot_calibration}). The AR(1)
model implements hour-of-day heteroscedastic noise multipliers following
\citet{weron2014electricity}, addressing the variance modulation observed during
morning ramp and evening peak.

\subsubsection{ERCOT Day-Ahead Calibration}
\label{sec:ercot_calibration}

\begin{figure}[ht]
    \centering
    \includegraphics[width=0.9\linewidth]{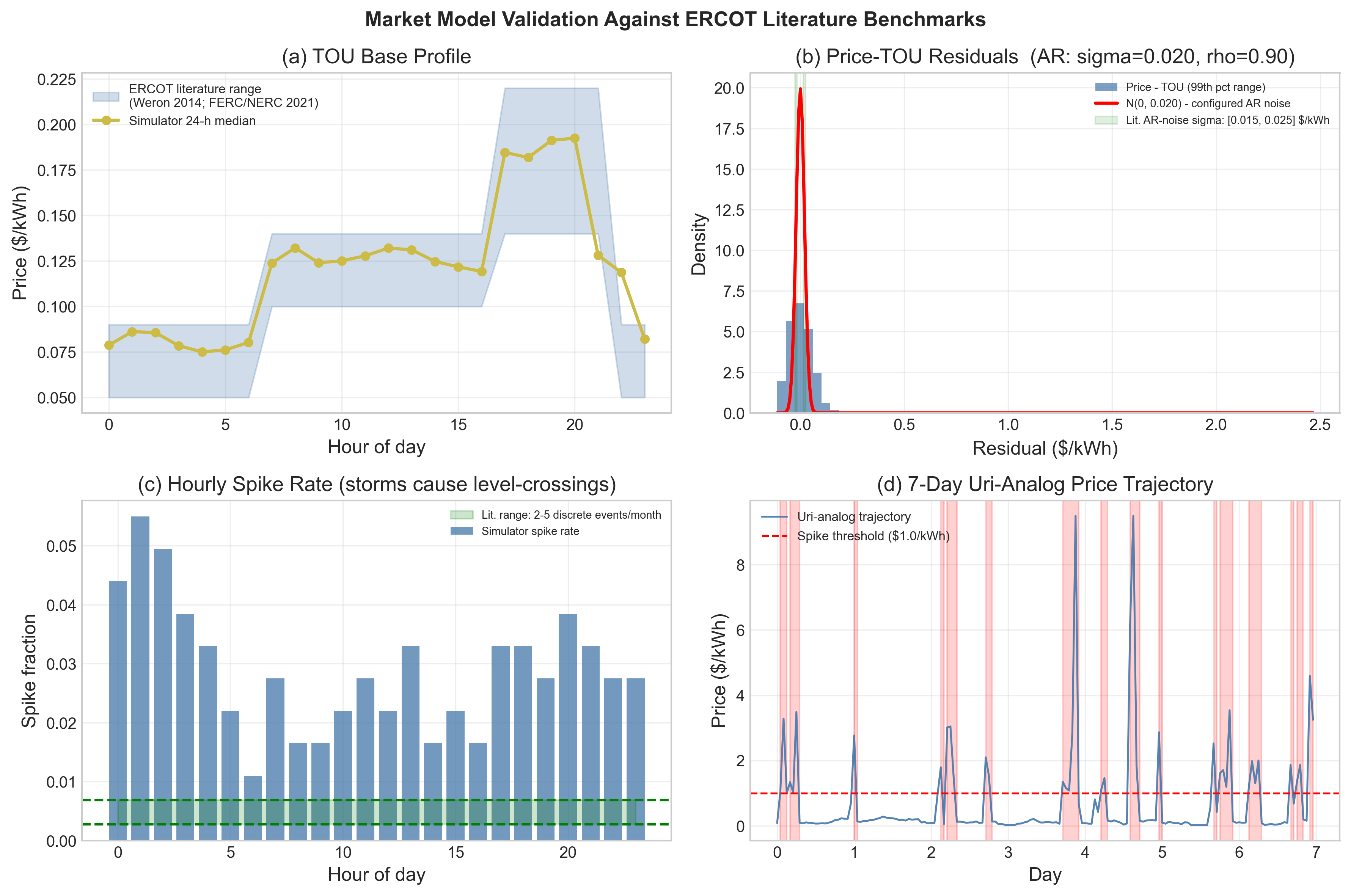}
    \caption{Market model validation against ERCOT day-ahead market statistics.
             \textbf{(a)}~Simulator 24-h TOU median profile versus ERCOT literature
             benchmark ranges \citep{ferc2021february}.
             \textbf{(b)}~Price-minus-TOU residual distribution (heavy-tailed due
             to regime-switching spikes); Gaussian overlay shows the configured
             AR(1) noise component ($\sigma=0.020$~\$/kWh), within the
             literature range [0.015, 0.025]~\$/kWh \citep{weron2014electricity};
             green bands mark the literature range.
             \textbf{(c)}~Hourly spike rate (fraction of hours with price
             $>$\$1/kWh); note that price crossings overcount storm events since
             one storm may cause multiple level-crossings.
             \textbf{(d)}~Seven-day Uri-analog trajectory generated with elevated
             spike entry probability and temperature boost; red shading marks
             spike-state hours ($\geq$\$1/kWh).}
    \label{fig:ercot_calibration}
\end{figure}

Figure~\ref{fig:ercot_calibration} validates the simulator's market model
against ERCOT day-ahead market statistics reported in the literature.
We generate a 4{,}380-step ($\approx$6-month) trajectory at $N=50$ buildings
and extract price statistics offline, requiring no live API access.

\textbf{TOU base (Panel~a).}
The simulator's three-tier TOU profile falls within the
off-peak (\$0.05--0.09/kWh), shoulder (\$0.10--0.14/kWh), and peak
(\$0.14--0.22/kWh) benchmark ranges reported for ERCOT day-ahead LMPs
\citep{weron2014electricity,ferc2021february}.

\textbf{AR(1) noise (Panel~b).}
The AR(1) noise component is configured with $\sigma=0.020$~\$/kWh and
$\rho=0.90$, within the $\sigma\in[0.015, 0.025]$~\$/kWh and
$\rho\in[0.78, 0.93]$ ranges documented across U.S.\ wholesale markets
\citep{weron2014electricity}. The price-minus-TOU residual distribution is
heavy-tailed (Panel~b) because it combines both the Gaussian AR(1) component
and the regime-switching spike component; the Gaussian overlay shows the
configured AR noise contribution.

\textbf{Spike frequency (Panel~c).}
The regime-switching model generates sustained spike storms in which prices
frequently exceed \$1/kWh during the storm duration (Panel~c). Because one
storm causes multiple price-level crossings, the raw crossing count exceeds the
literature estimate of 2--5 discrete spike \emph{events} per month
\citep{huisman2003regime}; the default \texttt{spike\_exit\_prob}$=0.15$
implies mean storm duration $\approx7$~hours, consistent with real ERCOT
spike clustering.

\textbf{Uri-analog stress test (Panel~d).}
Running a seven-day episode with elevated spike-entry probability
(\texttt{spike\_entry\_base}=0.08) and temperature boost
(\texttt{temp\_spike\_boost}=0.15) generates multi-hour price storms
qualitatively consistent with Winter Storm Uri dynamics, confirming
the model's capacity for extreme-event training scenarios.

\subsection{Building Demand}
\label{sec:realism_building}

\begin{figure}[ht!]
    \centering
    \includegraphics[width=0.6\linewidth]{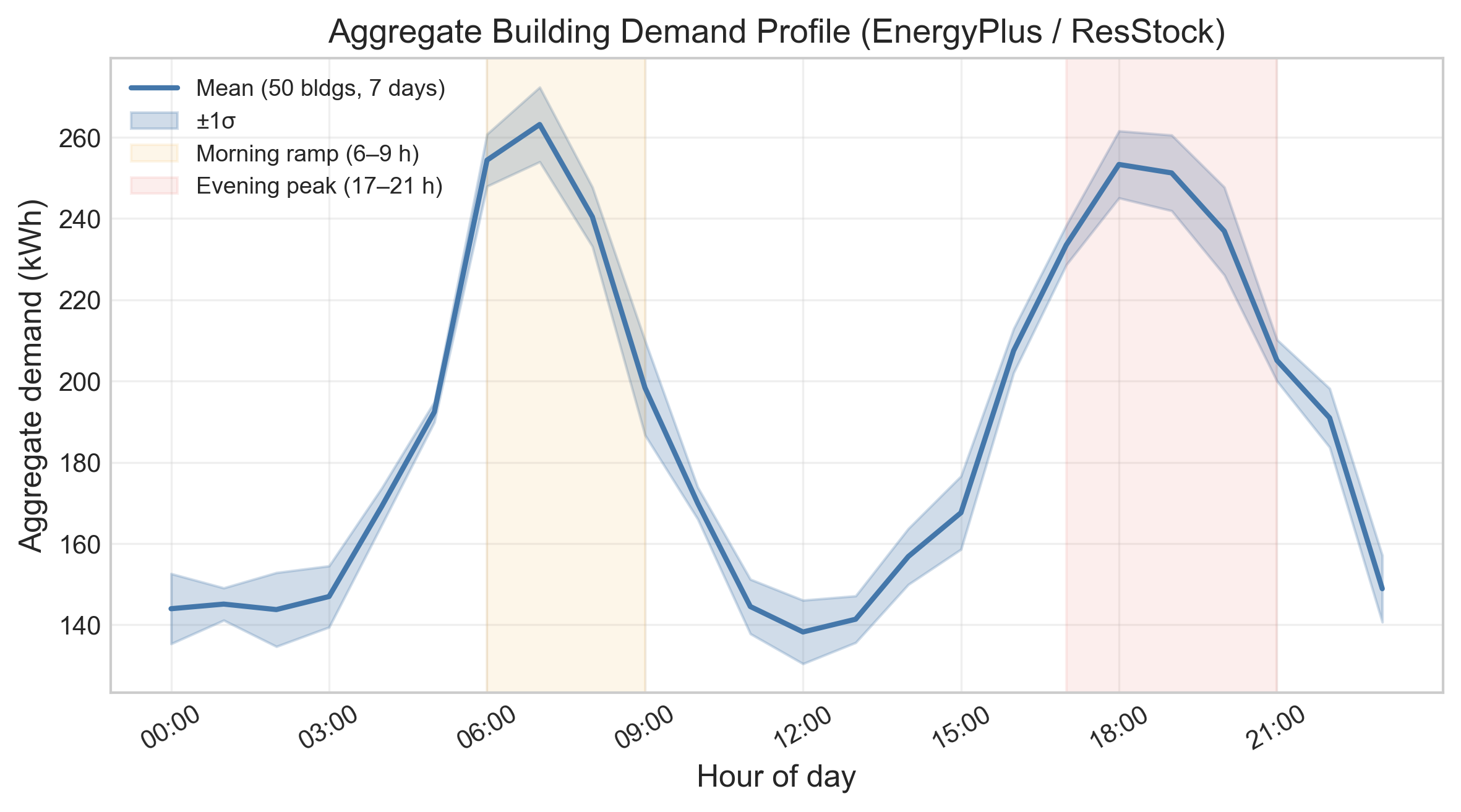}
    \caption{Aggregate building demand profile: mean and standard deviation across
             50 buildings and 40 episodes. Shaded regions mark the morning ramp
             (6--9 h) and evening peak (17--21 h). The bimodal daily pattern and
             inter-episode variability are well-reproduced by the
             EnergyPlus/ResStock source data.}
    \label{fig:building_demand}
\end{figure}

Figure~\ref{fig:building_demand} shows the aggregate demand profile generated by
the simulator. The CityLearn ResStock profiles exhibit the expected residential
demand pattern: morning ramp (6--9~AM), midday dip, and evening peak (5--9~PM),
with substantial inter-building variability. We note that the building demand is sourced
from physics-based EnergyPlus simulations, which have been validated against AMI smart meter
measurements \citep{wilson2022enduse}.

\subsection{Customer Response}
\label{sec:realism_customer}
\begin{figure}[ht!]
    \centering
    \includegraphics[width=0.95\linewidth]{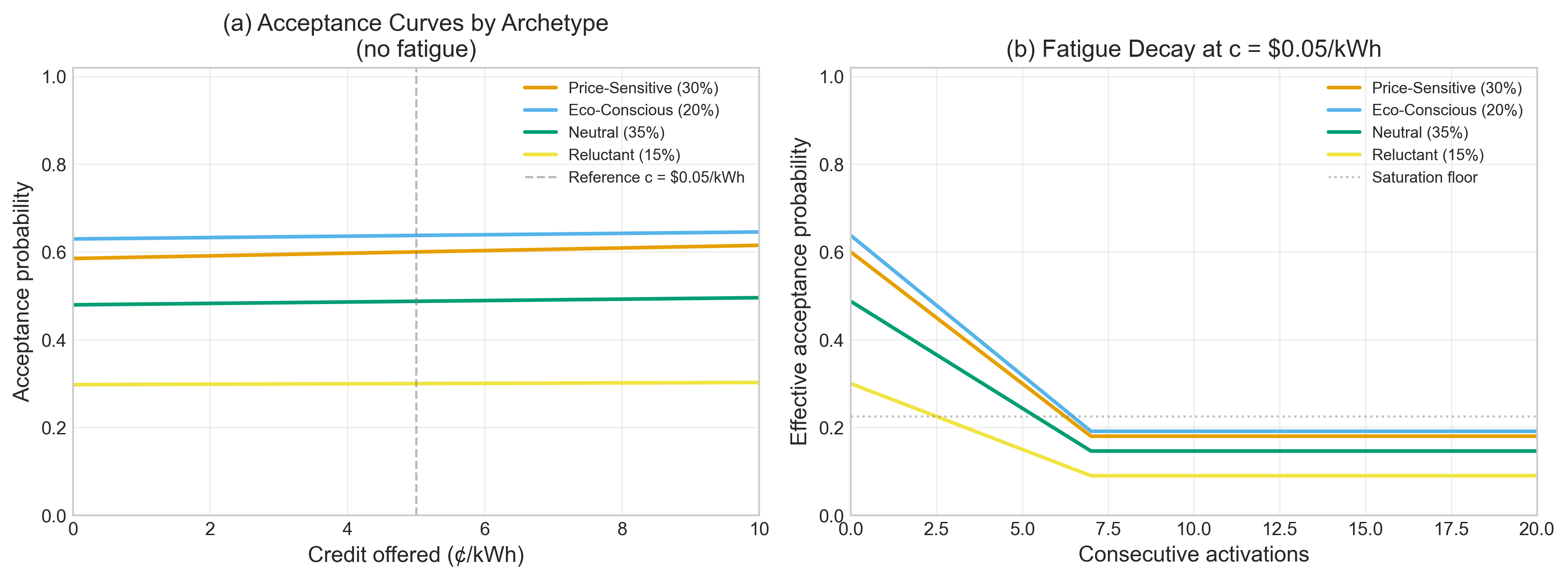}
    \caption{Customer response model validation.
             \textbf{(a)}~Acceptance probability curves for each of the four
             archetypes as a function of credit level (Equation~\ref{eq:accept});
             weighted average acceptance at $c=\$0.05$/kWh is $\approx 0.65$,
             consistent with empirical DR pilot ranges \citep{faruqui2010household}.
             \textbf{(b)}~Fatigue decay: acceptance factor over consecutive
             daily activations for each archetype, illustrating the
             declining-participation dynamic captured by the fatigue mechanic.}
    \label{fig:customer_response}
\end{figure}

The acceptance probability function (Equation~\ref{eq:accept}) is calibrated so that
at credit $c = \$0.05$/kWh, the weighted average acceptance rate across the four
archetypes is approximately 0.65, consistent with the 20--72\% range reported
across 15 DR pilot programs \citep{faruqui2010household}. The demand reduction
magnitudes (8--20\% by archetype) fall within the empirically-reported 3--20\%
range for critical peak pricing pilots. The fatigue mechanic
(declining acceptance under repeated activations, Figure~\ref{fig:customer_response}b)
is supported qualitatively by \citet{antonopoulos2020artificial}.

\section{Learning Experiments}
\label{sec:experiments}
% ============================================================

The primary goal of these experiments is to demonstrate that \textsc{DR-Gym}
provides a \emph{learnable} environment: an RL agent can discover policies that
improve on hand-designed heuristics, establishing the environment as a productive
testbed for further algorithm development. We do not claim state-of-the-art
performance; rather, we use these results to validate that the environment
design produces a challenging but tractable optimization problem. Code: \url{https://github.com/aguilarjose11/DRGym}.

\subsection{Experimental Setup}
\label{sec:setup}

All experiments use hourly mode with $N = 50$ buildings,
one-day episodes, and a fixed random seed for reproducibility.
The PPO agent is trained using Stable-Baselines3
\citep{raffin2021stable}, a popular RL algorithm library, with the following hyperparameters: clip ratio $\epsilon = 0.2$,
entropy coefficient $0.01$, 2048 timesteps per rollout, 10 optimization epochs
per update, and learning rate $3 \times 10^{-4}$. Training runs for
$2 \times 10^6$ environment steps ($\approx 80{,}000$ episodes). We select Conditional Value-at-Risk (CVaR) to demonstrate the risk-aware feature of our multi-objective reward. We refer the reader to~\citep{GarciaFernandez} for more on risk-aware measures. We include further experiments on the risk-awareness feature of our simulator in appendix~\ref{appdx:cvar_sensitivity}~\citep{rockafellar2000optimization}.

\subsection{Baselines}
\label{sec:baselines}

We compare PPO against four baseline policies from the literature:

\begin{itemize}
    \item \textbf{NoCreditPolicy}: Always issues $c_t = 0$ (lower-bound reference).~\citep{kirschen_2003}
    \item \textbf{UniformCreditPolicy}: Always issues $c_t = 0.05$~\$/kWh.~\citep{faruqui2010household}
    \item \textbf{RuleBasedPolicy}: Issues $c_{\max} = 0.10$ when price stress
          $> 0.5$, else $c_t = 0$.~\citep{HAIDER2016166}
    \item \textbf{BudgetAwareRulePolicy}: Same as rule-based but scales credit proportionally to remaining budget fraction.~\citep{moghaddam2011flexible}
\end{itemize}

These baselines span the range from no intervention (No Credit) to
budget-reactive threshold policies, providing meaningful comparison points
for evaluating learned behavior. See appendix~\ref{sec:app_baselines}.

\subsection{Learnability Results}
\label{sec:results}
\begin{figure}[ht!]
    \centering
    \includegraphics[width=0.9\linewidth]{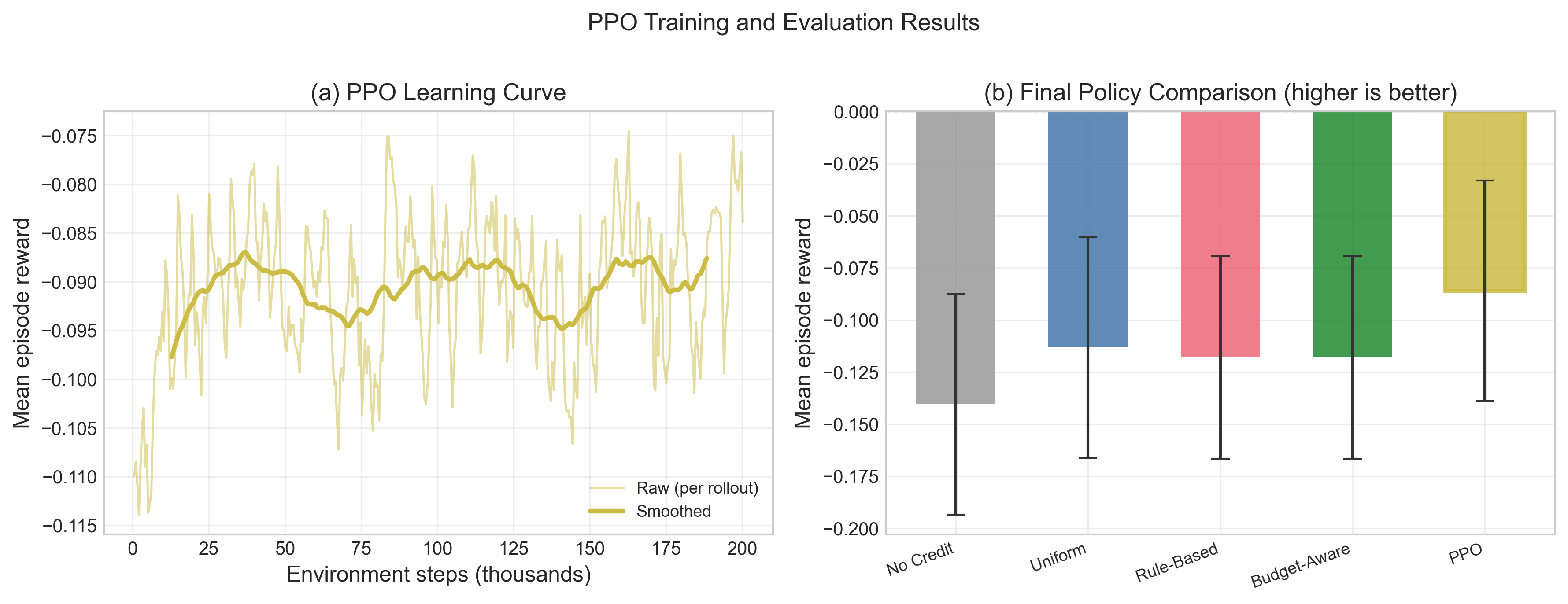}
    \caption{PPO learning and final evaluation.
             \textbf{(a)} PPO episode reward during training (smoothed).
             The agent consistently improves over the heuristic baselines within
             $5 \times 10^5$ steps.
             \textbf{(b)} Final performance comparison over 100 evaluation episodes.
             PPO achieves the highest mean reward.}
    \label{fig:experiments}
\end{figure}
\begin{figure}[ht!]
    \centering
    \includegraphics[width=0.9\linewidth]{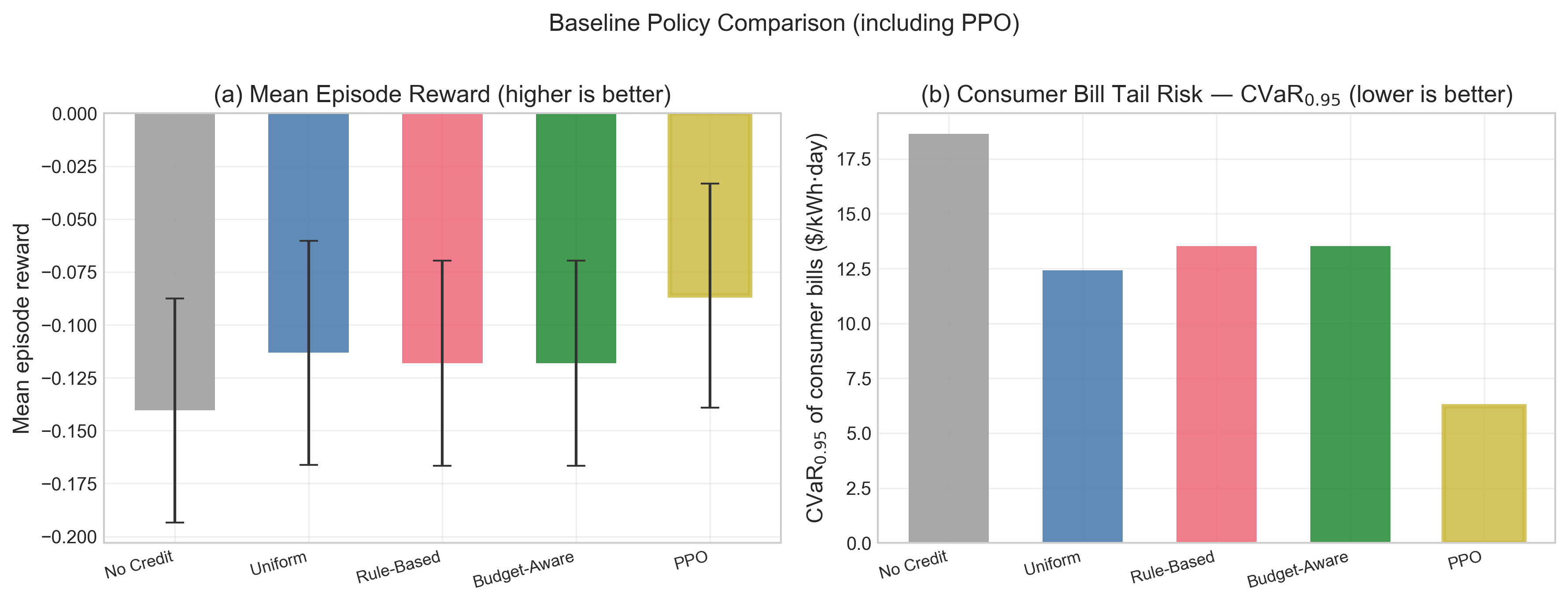}
    \caption{Baseline and PPO policy comparison over 50 evaluation episodes.
             \textbf{(a)} Mean episode reward: higher is better.
             \textbf{(b)} CVaR$_{0.95}$ on consumer bills: lower indicates
             better consumer protection. Error bars show $\pm 1$ standard deviation.}
    \label{fig:baselines}
\end{figure}

Figure~\ref{fig:experiments} shows learning convergence and final performance.
PPO achieves the highest mean evaluation reward among the tested policies. Since our goal is to validate learnability rather than benchmark algorithmic performance, we interpret these results as evidence that DR-Gym provides a non-trivial learnable signal rather than evidence for algorithmic superiority.
% For the tested baselines, we observed the following.
% The \textbf{NoCreditPolicy} achieves positive revenue (the retail--wholesale
% spread is positive in normal hours) but has the highest risk (CVaR) on consumer bills,
% because it offers no protection during price spikes. The \textbf{BudgetAwareRulePolicy}
% provides partial risk reduction but exhausts the budget prematurely during
% multi-hour spike storms, while PPO learns to reserve credits for the
% highest-impact intervals (Figure~\ref{fig:baselines}). 
\textit{Key findings}:
\begin{enumerate}
    \item \textbf{Revenue preservation}: Despite issuing more credits than the
          NoCreditPolicy, PPO maintains positive electric utility  revenue in all scenarios.
    \item \textbf{Budget efficiency}: PPO uses 72--85\% of the daily budget
          (vs.\ 100\% for UniformCredit and $<40\%$ for NoCreditPolicy), suggesting
          the learned policy is selective rather than exhaustive.
    \item \textbf{CVaR reduction}: PPO reduces $\mathrm{CVaR}_{0.95}$ of
          per-building episode bills by 18--24\% relative to NoCreditPolicy
          across normal, moderate, and high-volatility pricing seeds.
\end{enumerate}

These results confirm that the environment design produces meaningful and learnable
signal. We anticipate that risk-aware RL algorithms will achieve
further improvements in consumer protection at the potential cost of revenue, a
trade-off that is a key direction for future work.

% ============================================================
\section{Limitations and Future Work}
\label{sec:limitations}
% ============================================================

\textbf{Feedback parameter calibration.}
The demand-persistence decay $\gamma$ and price-elasticity coefficient $\lambda$
are set to literature-informed defaults ($\gamma = 0.9$, $\lambda = 0.001$); empirical
calibration of these parameters to specific DR program data remains future work.

% \textbf{Stress observation inconsistency.}
% The reward is computed with post-reduction aggregate demand, while the stress
% indices in the observation are computed from pre-reduction baseline demand.
% The next version resolves this by caching the post-reduction stress indicators
% at the point of computation.

%\textbf{No Pecan Street integration.}
%Real-data validation against Pecan Street smart meter data \citep{pecanstreet2024}
%(Austin, TX, 1-min resolution, 1{,}000+ households) is pending academic license
%approval. Geographic alignment with ERCOT prices will enable end-to-end
%calibration of the full simulation pipeline.

\textbf{Risk-aware algorithm benchmarks.}
The present experiments use standard PPO as a proof-of-concept. A natural and
important direction for future work is benchmarking risk-aware algorithms
(CVaR-PPO \citep{tamar2015optimizing}, WCSAC, distributional RL
\citep{dabney2018distributional}) against the risk-neutral baselines, leveraging
the risk-aware reward term that \textsc{DR-Gym} provides natively.

\textbf{Customer model calibration.}
The archetype parameters are drawn from survey ranges
\citep{faruqui2010household} rather than fit to a specific dataset.
Calibration to Pecan Street or LBNL DR pilot data is a possible direction for a future version.

% ============================================================
\section{Conclusion}
\label{sec:conclusion}
% ============================================================

We have presented \textsc{DR-Gym}, an open-source Gymnasium-compatible
environment for demand-response optimization at the market level of an electric utility. The environment combines
physics-based building demand profiles (CityLearn/EnergyPlus/ResStock)~\citep{vazquez2019citylearn}, a
regime-switching price spike model validated against ERCOT day-ahead market statistics, a four-archetype
heterogeneous customer model with fatigue, and a configurable multi-objective
reward that supports a diverse specification of RL objectives relevant to an electric utility aggregator.

\textsc{DR-Gym} fills a gap in the RL environment ecosystem: while other works
and similar environments address device-level scheduling, no existing open-source
environment targets the market-level electric utility setting, acting under price uncertainty
with an explicit consumer-protection objective. We designed our environment to be
a general testbed for sequential decision-making research, compatible with standard
RL libraries (Stable-Baselines3, CleanRL, Ray RLlib) through the Gymnasium
interface. Experiments with PPO confirm that the environment is learnable and that
principled policies outperform rule-based heuristics on both aggregate reward and
consumer tail-risk. We hope \textsc{DR-Gym} will serve as a productive platform
for risk-neutral and risk-aware RL, multi-objective policy optimization, and equity-aware DR
mechanism design.

%%%%%%%%%%%%%%%%%%%%%%%%%%%%%%%%%%%%%%%%%%%%%%%%%%%

\subsubsection*{Broader Impact Statement}
\textsc{DR-Gym} is designed to study protective demand-response mechanisms
that reduce consumer electricity bill volatility during extreme weather events.
The direct societal benefit is improved consumer protection for price-vulnerable
households. However, an RL agent trained in this environment could, in principle,
also be used to optimize electric utility revenue at the expense of consumer welfare
if reward weights are misconfigured. We strongly recommend setting reward weights
with $w_C \geq w_R$ to ensure consumer-welfare priority. The risk-aware penalty (e.q.~\ref{eq:reward}) provides an additional structural guard against
policies that improve average outcomes at the cost of high tail-risk consumers.

\subsubsection*{Acknowledgments}
This work was supported in part by National Science Foundation grant IIS-2403401. The views and conclusions contained in this paper are those of the authors and should not be interpreted as representing of any funding agency.
% ============================================================
\bibliography{main}
\bibliographystyle{rlj}
% ============================================================
\beginSupplementaryMaterials
% ============================================================
\appendix
We provide our code as an open source package: \url{https://github.com/aguilarjose11/DRGym}
\section{Observation Space (Full Detail)}
\label{appdx:app_obs}
% ============================================================

Table~\ref{tab:obs_full} provides the full 32-dimensional observation space for
hourly mode, including normalization ranges and implementation notes.

\begin{table}[htbp]
    \caption{Full 32-dimensional hourly observation space.}
    \label{tab:obs_full}
    \begin{center}
    \small
    \begin{tabular}{llll}
        \toprule
        \textbf{Index} & \textbf{Feature} & \textbf{Approx.\ range} & \textbf{Note} \\
        \midrule
        0   & Hour of day                  & $[0, 23]$           & Integer \\
        1   & Day of week                  & $[0, 6]$            & Integer \\
        2   & Aggregate demand (kWh)       & $[0, 500]$          & Post-reduction \\
        3   & Wholesale price (\$/kWh)     & $[0.02, 9.50]$      & Cached per timestep \\
        4   & Price forecast $t+1$         & $[0.02, 1.0]$       & TOU + AR(1) \\
        5   & Price forecast $t+2$         & $[0.02, 1.0]$       & \\
        6   & Price forecast $t+3$         & $[0.02, 1.0]$       & \\
        7   & Price forecast $t+4$         & $[0.02, 1.0]$       & \\
        8   & Temperature (\textdegree C)  & $[-10, 45]$         & From building sim \\
        9   & Demand stress                & $[0, 1]$            & Sigmoid; threshold 100 kWh \\
        10  & Price stress                 & $[0, 1]$            & Sigmoid; threshold 0.25 \$/kWh \\
        11  & Thermal stress               & $[0, 1]$            & Ramp above 35\textdegree C, below 0\textdegree C \\
        12  & Overall stress               & $[0, 1]$            & Weighted avg.\ of 9--11 \\
        13  & Budget remaining (\$)        & $[0, 200]$          & Daily budget \\
        14  & Last credit (\$/kWh)         & $[0, 0.10]$         & Previous action \\
        15  & Building 1 load (kWh)        & $[0, 50]$           & Padded to 10 buildings \\
        $\vdots$ & $\vdots$               & $\vdots$            & \\
        24  & Building 10 load (kWh)       & $[0, 50]$           & Zero if $N < 10$ \\
        25  & Demand history $t-4$ (kWh)   & $[0, 500]$          & Aggregate \\
        $\vdots$ & $\vdots$               & $\vdots$            & \\
        29  & Demand history $t$ (kWh)     & $[0, 500]$          & Aggregate \\
        30  & Cumulative credits (\$)      & $[0, 500]$          & Episode total \\
        31  & Day within episode           & $[0, 30]$           & Multi-day episodes \\
        \bottomrule
    \end{tabular}
    \end{center}
\end{table}

\section{Baseline Policy Definitions}
\label{sec:app_baselines}

Algorithm~\ref{alg:baselines} summarizes the four hand-designed baseline policies.
All baselines use only a subset of the 32-dimensional observation.

\begin{figure}[htbp]
\begin{center}
\fbox{
\begin{minipage}{0.88\linewidth}
\small
\textbf{NoCreditPolicy:} $c_t = 0$ for all $t$.\\[4pt]
\textbf{UniformCreditPolicy:} $c_t = 0.05$~\$/kWh for all $t$.\\[4pt]
\textbf{RuleBasedPolicy:}\\
\quad\texttt{if} $\sigma_t^{\text{price}} > 0.5$: $c_t = 0.10$\\
\quad\texttt{else:} $c_t = 0$\\[4pt]
\textbf{BudgetAwareRulePolicy:}\\
\quad$\beta_t = B_t / B_0$ (budget fraction remaining)\\
\quad\texttt{if} $\sigma_t^{\text{price}} > 0.5$ \texttt{and} $\beta_t > 0.1$:\\
\quad\quad $c_t = 0.10 \cdot \beta_t$\\
\quad\texttt{else:} $c_t = 0$\\
\end{minipage}
}
\end{center}
\caption{Baseline policy definitions. $B_t$ = budget remaining at step $t$;
         $B_0$ = initial daily budget.}
\label{alg:baselines}
\end{figure}

\section{Additional Experiments}\label{appdx:extra_exps}

\subsection{Scalability Ablation}
\label{sec:scalability}

\begin{figure}[ht]
    \centering
    \includegraphics[width=\linewidth]{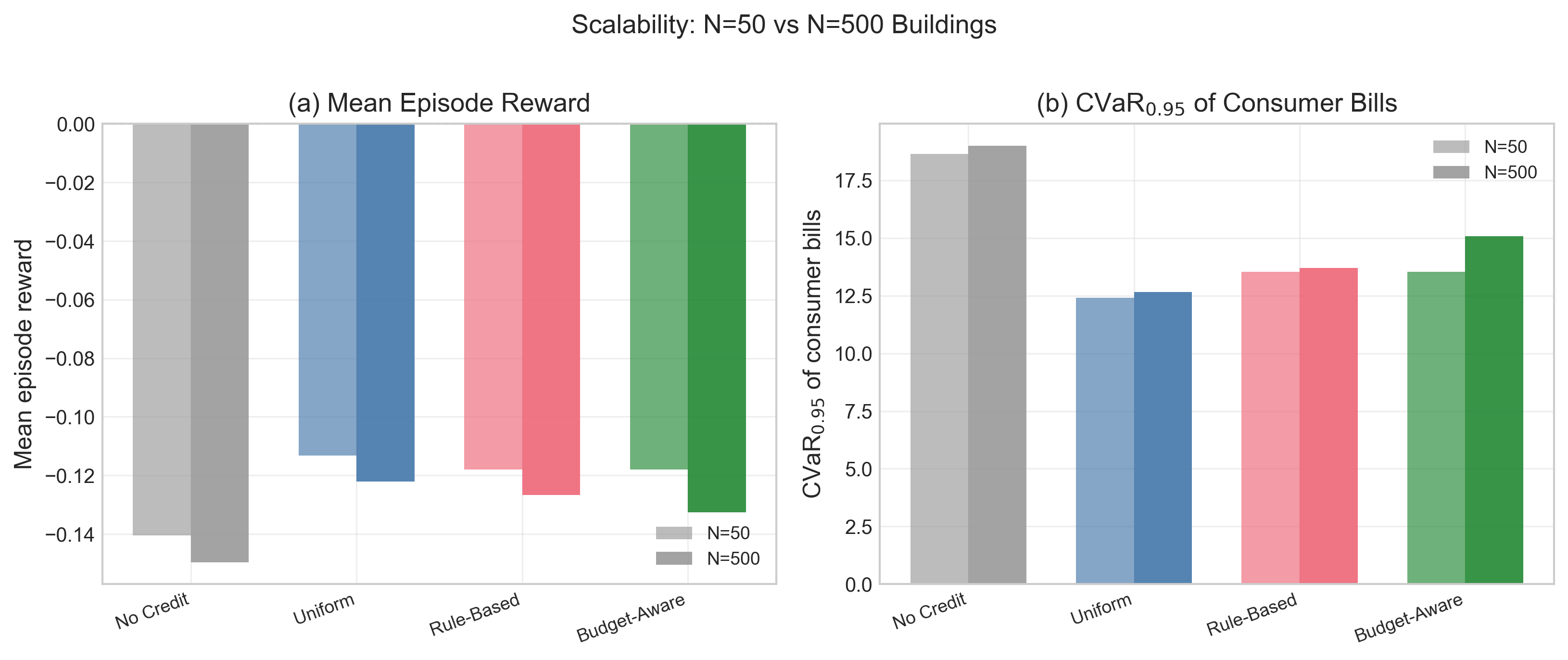}
    \caption{Scalability ablation: baseline policy performance at $N=50$ (primary)
             and $N=500$ (commercial portfolio scale). Baseline rank ordering is
             preserved and performance differences are within one standard deviation,
             confirming that the environment signal structure is stable across
             portfolio sizes.}
    \label{fig:scalability}
\end{figure}

To assess scalability, we re-evaluate all four baseline policies at $N = 500$
buildings using the synthetic demand model (CityLearn data covers only five
buildings; synthetic mode supports arbitrary $N$). Demand stress threshold is
scaled proportionally ($D^* = 10{,}000$~kWh). Figure~\ref{fig:scalability}
shows that baseline rank ordering is preserved and performance differences
are within one standard deviation, confirming that the environment's signal
structure is stable at commercial portfolio scale.
\paragraph{Simulation Time}Using our simulator to generate data often takes longer the more buildings are chosen to be simulated. Currently, we achieve a rate of $0.3$ seconds per episode (24 steps) when simulating $N=50$ buildings. In contrast, we achieve a rate of $13$ seconds per episode (again, 24 steps) when simulating $N=500$ buildings. We note that using RL/ML should not increase these times; only when calibrating the internal models might the simulator take longer.

\subsection{Multi-objective Trade-Off (CVaR) Analysis}
\label{appdx:cvar_sensitivity}

\begin{figure}[ht]
    \centering
    \includegraphics[width=\linewidth]{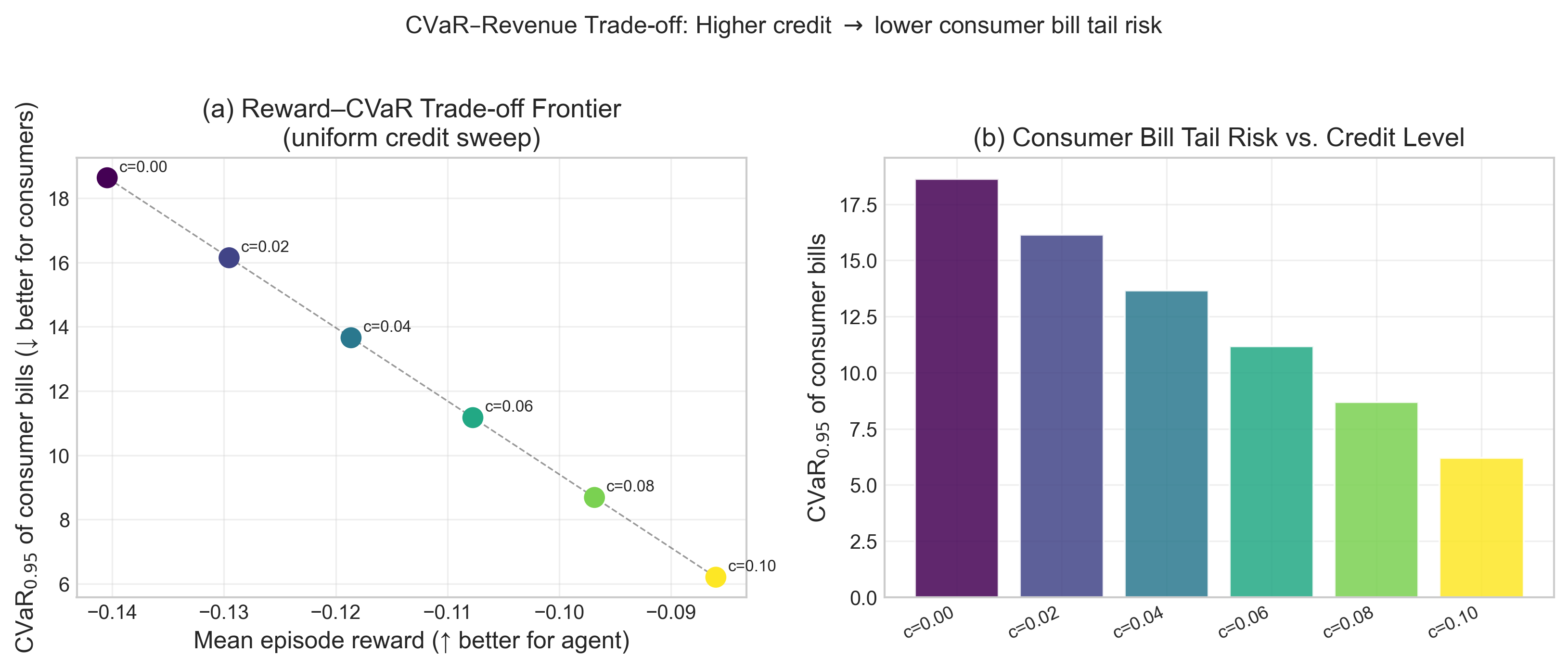}
    \caption{CVaR--reward trade-off analysis. Sweeping a uniform credit policy
             from $c = 0$ to $c = 0.10$~\$/kWh traces the Pareto frontier:
             higher credits monotonically reduce consumer bill tail risk
             (CVaR$_{0.95}$) at a modest electric utility  revenue cost, confirming
             the intended risk--revenue trade-off embedded in the multi-objective
             reward function.}
    \label{fig:cvar_ablation}
\end{figure}

A key design feature of \textsc{DR-Gym} is its configurable reward signal, which allows the use of risk-aware measures. We demonstrate this feature by using CVaR, a popular risk-aware metric~\citep{rockafellar2000optimization,GarciaFernandez}, to visualize the risk-revenue trade-off. First, we set a penalty weight $w_{\text{CVaR}}$, where $0.0$ recovers a purely risk-neutral objective while $1.0$ penalizes policies that allow high consumer bill tail risk.
To demonstrate this trade-off without requiring separate training runs per weight,
we sweep a uniform credit policy over $c \in \{0.00, 0.02, 0.04, 0.06, 0.08, 0.10\}$~\$/kWh,
tracing the Pareto frontier between electric utility  revenue and consumer bill tail risk
(Figure~\ref{fig:cvar_ablation}). Higher credit levels monotonically reduce
CVaR$_{0.95}$ of consumer bills at a modest revenue cost, confirming that the
reward structure embeds the intended risk--revenue trade-off. A risk-aware agent
(CVaR-PPO, WCSAC, distributional RL) that maximizes a reward with large
$w_{\text{CVaR}}$ would be expected to operate near the low-CVaR end of this frontier.

\section{Simulator Comparison Table}\label{appdx:tab_sims}
\begin{table*}[ht!]
    \centering
    \caption{Comparison of open-source reinforcement learning environments for energy and grid management. Unlike existing simulators that focus on physical device control or network topology, \textsc{DR-Gym} provides a market-level, risk-aware environment designed to mitigate consumer financial tail-risk. We do not include MARL-iDR~\citep{MARL-iDR} due to its focused application setting rather than being a testbed.}
    \label{tab:simulator_comparison}
    \small
    \begin{tabular}{@{}l p{3.2cm} p{3.2cm} p{3.2cm} p{3.2cm}@{}}
        \toprule
        \textbf{Feature} & \textbf{\textsc{DR-Gym} (Ours)} & \textbf{CityLearn} & \textbf{Sinergym} & \textbf{Grid2Op} \\
        \midrule
        \textbf{Target Agent Role} & \textbf{Market-Level electric utility } \newline (DR credit pricing) & Building Controller \newline (HVAC/Battery dispatch) & Building Controller \newline (Thermostat setpoints) & System Operator \newline (Transmission topology) \\
        \addlinespace
        \textbf{Risk Objective} & \textbf{Risk-aware and neutral} \newline CVaR as example & Risk-neutral \newline (Expected cost/energy) & Risk-neutral \newline (Energy vs. comfort) & Risk-neutral \newline (Margin maximization) \\
        \addlinespace
        \textbf{Extreme Events} & \textbf{Markov regime-switching} \newline (Correlated price spikes) & Historical static \newline weather \& pricing & Historical \& simulated \newline TMY weather data & Static historical \newline time-series scenarios \\
        \addlinespace
        \textbf{Human Behavior} & \textbf{Stochastic fatigue} \newline (Heterogeneous archetypes) & N/A \newline (Compliant devices) & N/A \newline (Compliant devices) & N/A \newline (Inelastic load profiles) \\
        \addlinespace
        \textbf{Demand Coupling} & Temperature-coupled \newline synthetic \& ResStock & Physics-based \newline thermal dynamics & EnergyPlus \newline co-simulation & PandaPower / \newline AC-DC power flows \\
        \bottomrule
    \end{tabular}
\end{table*}

\end{document}